\documentclass[9pt,twocolumn,twoside]{osajnl}
\usepackage{times}
\usepackage{amssymb}
\usepackage{multirow}
\usepackage{mathtools}
\usepackage{subfloat}
\usepackage{bbm}
\usepackage{algorithm}
\usepackage{algpseudocode}
\usepackage{pbox}
\usepackage{placeins}
\journal{ol} 
\setboolean{shortarticle}{false}

\usepackage{times}
\usepackage{graphicx}
\usepackage{subcaption}
\usepackage{units}
\usepackage{url}

\usepackage{amsmath}

\usepackage[printonlyused]{acronym}

\acrodef{RL}{reinforcement learning}
\acrodef{TO}{trajectory optimization}
\acrodef{PPO}{Proximal Policy Optimization}
\acrodef{MLP}{multilayer perceptron}
\acrodef{MPC}{model predictive control}
\acrodef{SOTA}{state-of-the-art}
\acrodef{TAMOLS}{terrain-aware motion generation for legged systems}
\acrodef{DTC}{Deep Tracking Control}
\acrodef{IK}{inverse kinematics}
\acrodef{WBC}{whole-body control}

\ifx\isReview\undefined
    \newcommand\highlight[1]{#1}
\else
    \usepackage{xcolor}
    \newcommand\highlight[1]{\textcolor{red}{#1}}
\fi

\title{Attention-Based Map Encoding for Learning Generalized Legged Locomotion} 

\author{
Junzhe~He$^{1}$, 
Chong~Zhang$^{1}$, 
Fabian~Jenelten$^{1}$,
Ruben~Grandia$^{2}$,
Moritz~B{\"a}cher$^{2}$,
Marco~Hutter$^{1}$ 
\\
\normalsize{$^{1}$Robotic Systems Lab, ETH Zurich, 8092 Zurich, Switzerland.} \\
\normalsize{$^{2}$Disney Research Zurich, Stampfenbachstrasse 48, 8006 Zurich, Switzerland.} \\
}

\begin{abstract}
Dynamic locomotion of legged robots is a critical yet challenging topic in expanding the operational range of mobile robots. It requires precise planning when possible footholds are sparse, robustness against uncertainties and disturbances, and generalizability across diverse terrains. While traditional model-based controllers excel at planning on complex terrains, they struggle with real-world uncertainties. Learning-based controllers offer robustness to such uncertainties but often lack precision on terrains with sparse steppable areas. Hybrid methods achieve enhanced robustness on sparse terrains by combining both methods but are computationally demanding and constrained by the inherent limitations of model-based planners.
To achieve generalized legged locomotion on diverse terrains while preserving the robustness of learning-based controllers, this paper proposes to learn an attention-based map encoding conditioned on robot proprioception, which is trained as part of the end-to-end controller using reinforcement learning. We show that the network learns to focus on steppable areas for future footholds when the robot dynamically navigates diverse and challenging terrains. We synthesize behaviors that exhibit robustness against uncertainties while enabling precise and agile traversal of sparse terrains. Additionally, our method offers a way to interpret the topographical perception of a neural network.
We have trained two controllers for a 12-DoF quadrupedal robot and a 23-DoF humanoid robot respectively and tested the resulting controllers in the real world under various challenging indoor and outdoor scenarios, including ones unseen during training.

\end{abstract}

\begin{document} 
\maketitle

\section*{Introduction}
\label{sec:intro}
Humans and legged animals inhabit almost every corner of our planet, adept at traversing various terrains in the wild.
Similarly, legged robots hold vast potential for navigating complex natural landscapes that are typically inaccessible to other wheeled or tracked mobile robots. 
Yet, navigating challenging terrains demands \textbf{1)} precise planning when possible footholds are sparse (e.g., on construction debris~\cite{jenelten2023dtc}), and \textbf{2)} robustness in the presence of uncertainties and disturbances~\cite{Lee2020,Miki2022}. Furthermore, the locomotion controller \textbf{3)} must be able to generalize across diverse terrains.

Toward this goal, various methods have been explored, including learning-based ones such as deep reinforcement learning (DRL), model-based ones such as model predictive control (MPC), and hybrid approaches that combine both. However, achieving generalized legged locomotion across diverse terrains with both precision and robustness remains an open problem. In this paper, we offer an attention-based learning framework to train robust and generalized controllers that can precisely navigate on various terrains.

DRL has emerged as a powerful tool for enabling robust and agile legged locomotion on challenging terrains. By training an actuator model through supervised learning and appropriately randomizing the training environment, Hwangbo et al.~\cite{Hwangbo2019} successfully transferred the dynamic motions learned through trial and error in simulation to the real world.
Lee et al.~\cite{Lee2020} and Siekmann et al.~\cite{cassie_stairs} developed robust learning-based controllers for blind traversal of rough terrains and stairs by quadrupedal and bipedal robots, respectively. 
Regarding perceptive locomotion, Rudin et al.~\cite{Rudin2021} trained end-to-end controllers on uneven terrains by massively parallel DRL. Miki et al.~\cite{Miki2022} further advanced robust perceptive quadrupedal locomotion in the wild by learning to filter out unreliable perception using a history of proprioceptive data. Additionally, DRL approaches have enabled legged robots to perform dynamic parkour maneuvers across various structured terrains~\cite{Rudin2022, anymal_parkour, zhuang2023robot, cheng2023parkour, zhuang2024humanoid, zhang2024wococo}. 
However, these approaches struggle on sparse terrains because it is hard for the DRL algorithm to discover valid footholds and learn from them. To tackle this problem, recent DRL-based work has achieved locomotion on sparse terrains through curriculum tuning to guide the training process with human intuition~\cite{chong_iros}, but it only overfits a small range of terrains with a single multilayer perceptron (MLP) policy network and cannot generalize. Another work enables a bipedal robot to walk on fake stepping stones (QR-code tags on the flat ground) through MLP-based foothold feasibility prediction~\cite{Duan2022LearningDB} but has not yet demonstrated locomotion on real terrains or generalization across various terrains.

Alternative to RL, model-based methods have been explored for decades to produce versatile movements and smooth trajectories that adhere to the kinematic and dynamic constraints of the robot.
Some work \cite{little_dog, Fankhauser2018} have enabled stable locomotion on rough terrains such as stairs, steps, inclines, etc. To achieve agility on uneven terrains or with various payloads, more adaptable frameworks such as MPC are used for locomotion tasks~\cite{6095035, Neunert2016FastNM,  Farshidian2017RealtimeMP, 8594448, SRBD_MPC} by solving optimal control problems over a long horizon. More recent endeavors have demonstrated stronger traversal performance on the real robot, achieving real-time versatile legged locomotion on uneven and sparse terrains by predicting accurate footholds~\cite{Mastalli2022, Ruben2022, Jenelten2022}. 
Despite this progress, model-based controllers remain susceptible to various assumptions introduced during the modeling process, such as perfect state estimation, perfect and complete map information, and simplified dynamic and kinematic models, which may lead to degraded performance under model-mismatch, drifted state estimation, and imprecise motor actuation, etc. 

To combine the advantages of both model- and learning-based controllers, hybrid methods have been proposed. Kwon et al.~\cite{Kwon2020} propose to train a dynamic model for model-based controllers. Some works~\cite{towr_learning, Surovik2021, Melon2021} warm-start non-linear solvers from learned initialization to speed up convergence. Other works utilize DRL to generate footholds that are then tracked by model-based controllers~\cite{RLOC, glide}. These approaches leverage learning to improve or bootstrap MPC while still running an MPC as the main controller and hence remain fragile to the uncertainty faced in real-world deployments. In contrast, to combine generalization capability and precise footholds predictions from a model-based planner and the robustness from a learning-based controller, DTC~\cite{jenelten2023dtc} trains a DRL controller to track the reference state trajectories generated by model-based optimization~\cite{Jenelten2022}, effectively overcoming model mismatch, slippage, and deformable terrains. Nonetheless, compared to end-to-end approaches, which directly map observations to actions, these hybrid controllers suffer from the complexity of the hierarchical structure. For example, during training, DTC requires running the model-based planner on a CPU and the training pipeline on a GPU and takes 14 days of training for full convergence. Furthermore, DTC requires running the MPC controller during deployment, which is computationally demanding and relies on the performance of the model-based planner - the height map is fed into the model-based controller, which might generate infeasible guidance under degraded perception.

In summary, an end-to-end learning-based approach that can achieve precise, robust, and generalized locomotion on sparse terrains is missing. Additionally, despite all the advancements in legged locomotion under different task specifications, no approach has yet demonstrated the successful sim-to-real transfer of perceptive controllers to achieve dynamic locomotion on sparse terrains for both quadrupedal and bipedal robots. 

In this paper, we propose to train an attention-based map encoding for generalized legged locomotion, which consists of two levels: \textbf{1)} a convolutional neural network (CNN) that embeds point-wise local terrain features in a robot-centric height map sampled from elevation mapping~\cite{elevationmapping_cupy}, and \textbf{2)} a multi-head attention (MHA)~\cite{NIPS2017_3f5ee243} module that queries point-wise map features and combines them with proprioceptive observations. MHA is a neural representation mechanism utilized in transformers~\cite{NIPS2017_3f5ee243}, a deep learning architecture that enables highly effective handling of sequential and multi-modal information. Recent works on legged locomotion learning have also introduced the transformer architecture to embed multi-modal observations~\cite{yang2022learning} or capture time-sequential features~\cite{jitendra_SR}, achieving locomotion on unstructured terrains such as grasslands, slopes, and flat ground. Instead of using a complete transformer model, we investigate how MHA can enable generalized legged locomotion across diverse terrains by focusing on the areas that matter most. In our framework, a low-level CNN learns effective extraction of local features for various terrains, and the MHA module selects useful terrain points based on the attention conditioned on proprioception, leading to a compact and generalizable representation of high-dimensional observations. We observe that the selected points indicate the future footholds without any supervised learning. 

To effectively synthesize our locomotion policies, we design a two-stage training pipeline. In the first stage, we train the controller on base terrains (defined in the supplementary methods, section "Training Details - Terrains") with perfect perception. This stage is crucial for initializing the map encoding learning and leads to basic locomotion skills under ideal conditions. In the second stage, we introduce more challenging terrains (defined in the supplementary methods, section "Training Details - Terrains") with perception noise and drift to enhance the generalization to real-world conditions.

\begin{figure*}
\centering
    \includegraphics[width = 7.3in]{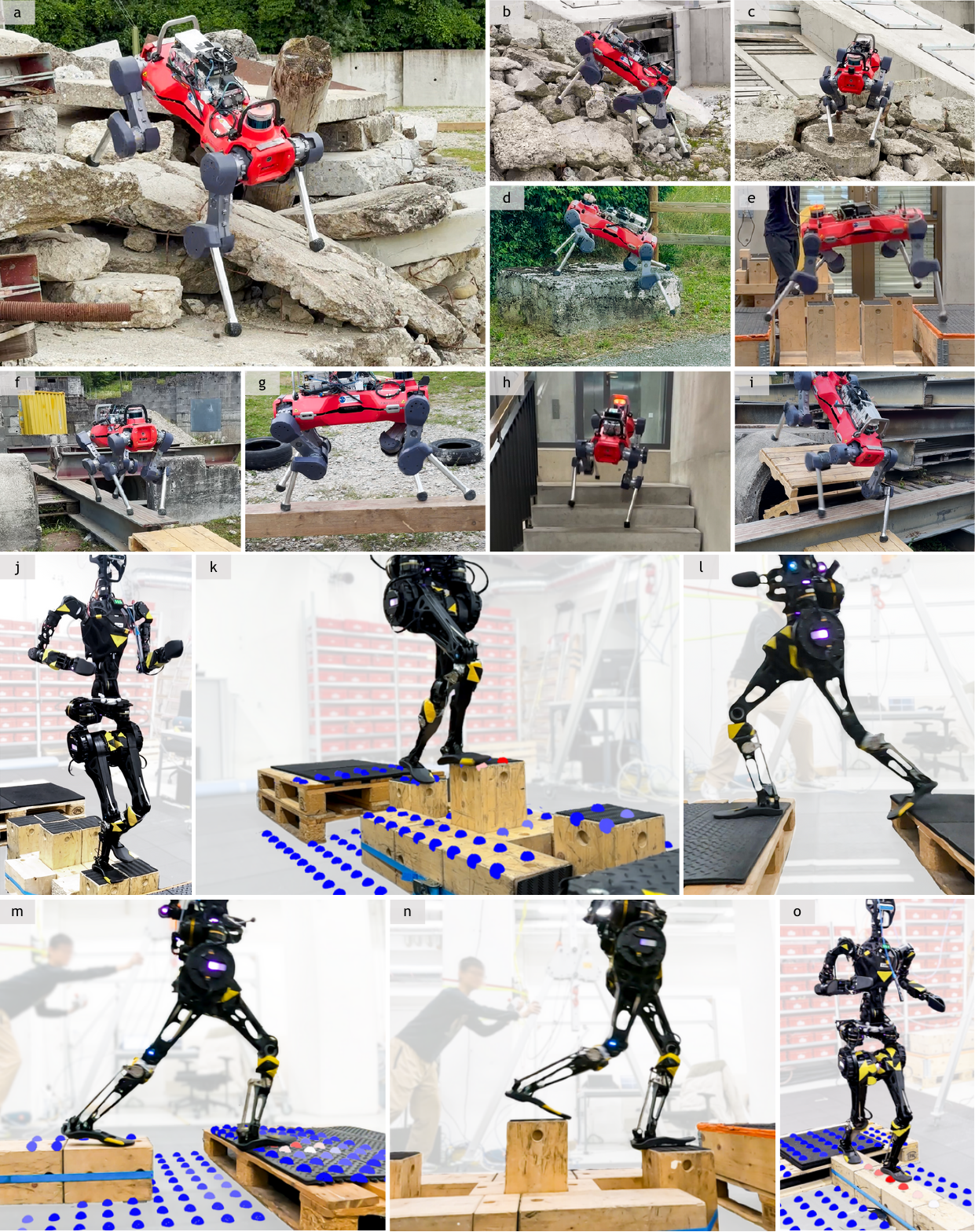}
    \caption{\textbf{\highlight{Learning interpretable, generalizable, agile, and robust legged locomotion on diverse terrains.}} Our end-to-end controllers enabled ANYmal-D (a - i) and GR-1 (j - o) to dynamically traverse diverse challenging terrains. Highly interpretable point-wise map encodings are graphically visualized in k, m, and o, where colors more intense in red represent higher attention weights, indicating the next foothold.}

    \label{pics:combined}
\end{figure*}

We demonstrate that our proposed end-to-end control framework has resulted in substantial advancements over the state-of-the-art (SOTA) in \textbf{1)} precise and generalized dynamic locomotion on a broad range of terrains~\cite{jenelten2023dtc} while achieving \textbf{2)} robustness against uncertainties and model mismatch, and producing \textbf{3)} an interpretable representation of map scans that can be graphically visualized, with \textbf{4)} the same framework being applicable to both a quadrupedal robot ANYmal-D~\cite{Hutter2016} and a humanoid robot Fourier GR-1~\cite{fourier_gr1_2023}, as shown in Figure~\ref{pics:combined}. 

\section*{Results}
\subsection{Precise and Generalized Locomotion}

\begin{figure*}
\centering
\includegraphics[width=7.3in]{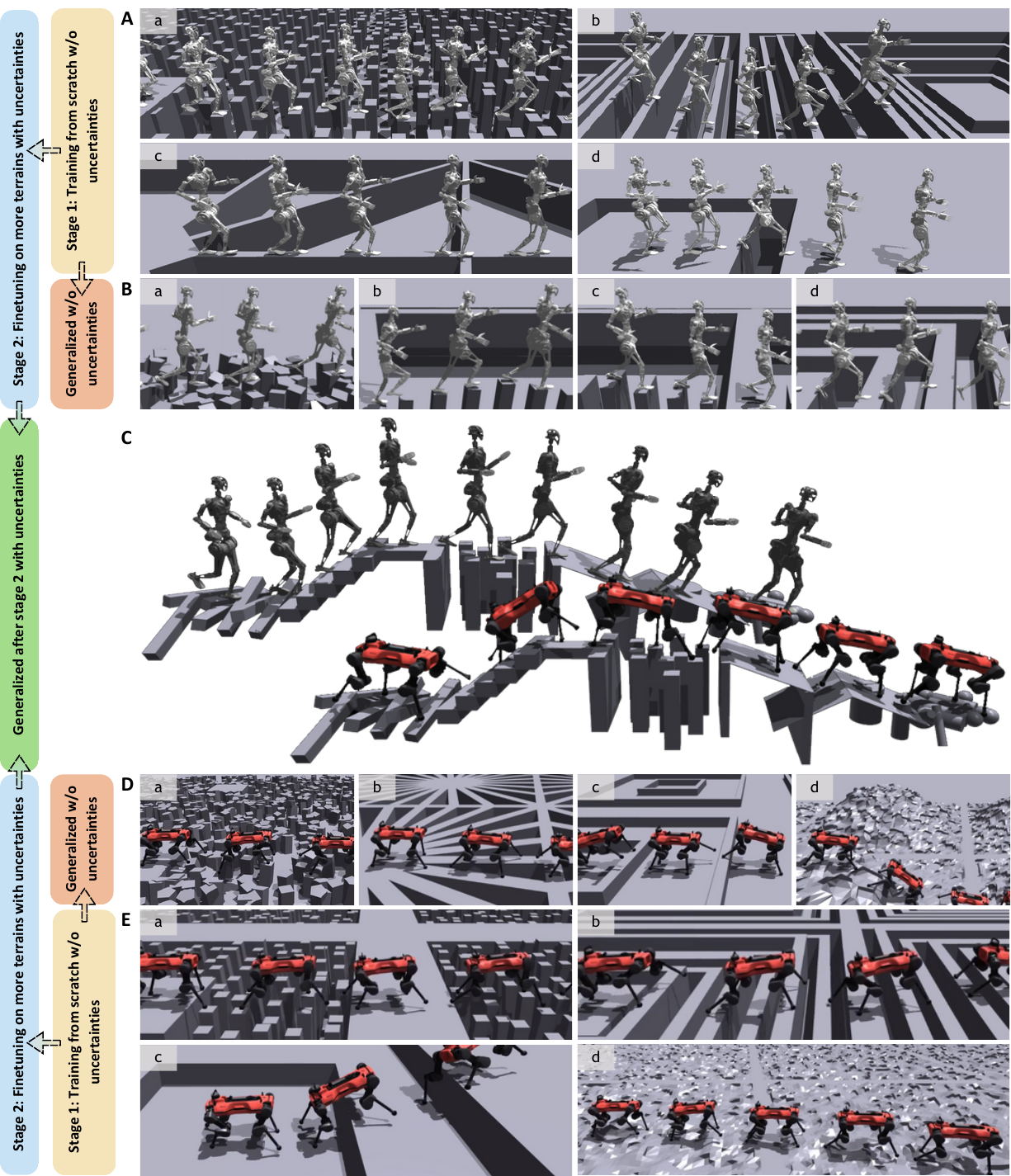}
\caption{\textbf{\highlight{Precise and Generalized Locomotion in Simulation.}} \textbf{(A)} Selection of base terrains for GR1 stage 1 training, including grid stones (a), pallets (b), beams (c), and gaps (d) \textbf{(B)} Selection of fine-tuning terrains for GR1 stage 2 training, including pentagon stones (a), single-column stones (b), narrow pallets (c), and consecutive gaps (d) \textbf{(C)} GR1 and ANYmal-D on obstacle parkour~\cite{Ruben2022}. \textbf{(D)} Selection of fine-tuning terrains for ANYmal-D stage 2 training, including pentagon stones (a), beams (b), rings (c), and rough hills (d). \textbf{(E)} Selection of base terrains for ANYmal-D stage 1 training, including grid stones (a), pallets (b), pits (c), and rough ground (d). }
\label{fig:generalization}
\end{figure*}

\begin{figure*}[h]
\centering
\includegraphics[width=7.3in]{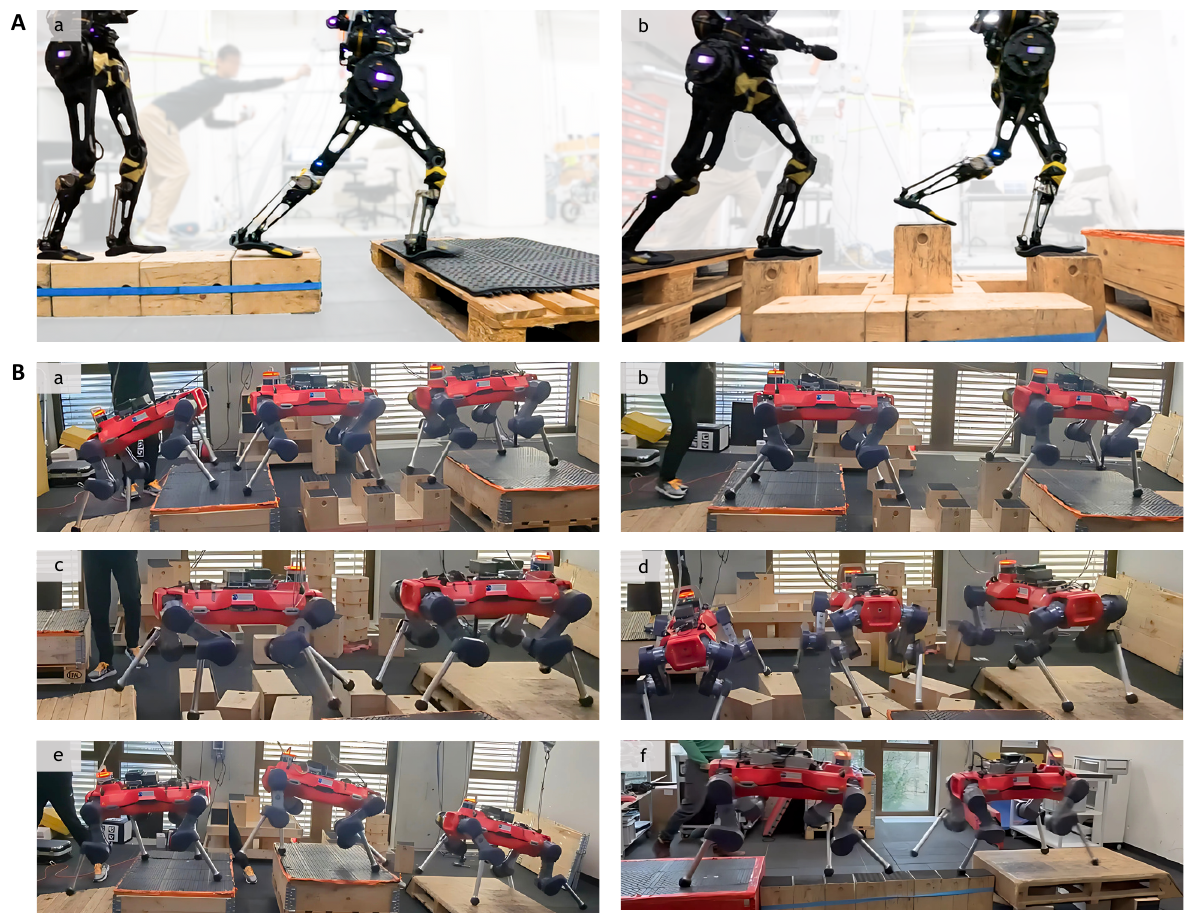}
\caption{\textbf{\highlight{Precise and Generalized Locomotion on Hardware.}} \textbf{(A)} Selection of sparse terrains for GR1, including beam and gap (a) and single-column stones with height difference (b). \textbf{(B)} Selection of sparse terrains for ANYmal-D, including stepping stones (a), stepping stones with height difference (b), randomly placed stepping stones forward (c), randomly placed stepping stones sideway (d), boxes and gaps (e), and a 19-cm wide beam (f). }
\label{fig:results_real}
\end{figure*}

We tested our learned policies on both the GR-1 and ANYmal-D robots in simulation (Figure~\ref{fig:generalization}) and extensive real-world experiments (Figure~\ref{pics:combined} and Figure~\ref{fig:results_real}) across a diverse range of terrains, including ones never encountered during the training phase. 
Figure~\ref{fig:generalization} A illustrates the controller’s performance on GR-1, trained from scratch during stage 1, where it exhibited \textbf{precise} foot placements across various terrain types, such as grid stones, pallets, beams, and gaps. Despite being exclusively trained on base terrain types partially shown in Figure \ref{fig:generalization} A, the controller demonstrated its ability to \textbf{generalize} to unseen terrains, including pentagon stones, single-column stones, narrow pallets, and consecutive gaps, shown in Figure~\ref{fig:generalization} B. This highlights our controller's capability to transfer the learned behaviors to new terrains out of the training distribution.

A similar pattern of generalization was observed in the ANYmal-D controller. The controller was able to adapt and perform effectively on unseen terrains, as shown in Figure~\ref{fig:generalization} D, despite being trained only on base terrains shown in Figure~\ref{fig:generalization} E (Note that the kinematics of ANYmal-D differs from GR-1, requiring adjustments in the terrain selection during training. More terrain details can be found in the supplementary methods, section "Training Details - Terrains")). This demonstrates the adaptability of our approach to managing different embodiments with varying kinodynamics, enabling generalization capabilities despite hardware variations.

To further enhance the policies’ \textbf{precision} on a wider range of terrains and \textbf{robustness} and in real-world experiments, we fine-tuned them in stage 2 by incorporating additional terrain types and introducing disturbances and uncertainties. This fine-tuning process improved the controller's ability to navigate more complex and unpredictable environments, reinforcing its stability and adaptability under real-world conditions. We demonstrate the resulting controllers in Figure~\ref{fig:generalization} C, where both GR-1 and ANYmal-D achieve a 100\% success rate in traversing a challenging obstacle parkour (with disturbances and uncertainties) designed by Grandia et al.~\cite{Ruben2022} that had not been encountered during training. Notably, we are the first to enable dynamic humanoid locomotion across such mixed sparse terrains with an online controller. The same controllers also achieve high robustness in real-world deployments, demonstrated in Figure~\ref{pics:combined} and Figure~\ref{fig:results_real}, unifying the properties of SOTA model-based and learning-based approaches.

We deployed the fine-tuned policies for both ANYmal-D and GR-1 zero-shot on the real hardware. To validate the controller's precision and robustness, we tested on various sparse terrains, including gaps, stepping stones, and beams, as depicted in Figure \ref{fig:results_real}. Our results represent advances in robotic locomotion, demonstrating that our end-to-end DRL-based controller can achieve precise and generalized performance across a wide range of challenging terrains. 
Prior to our work, no other end-to-end DRL-based controllers have achieved such a level of generalization.

\subsection{Agility and Recovery Reflexes by Whole-Body Coordination}

\begin{figure*}
\centering
\includegraphics[width=7.3in]{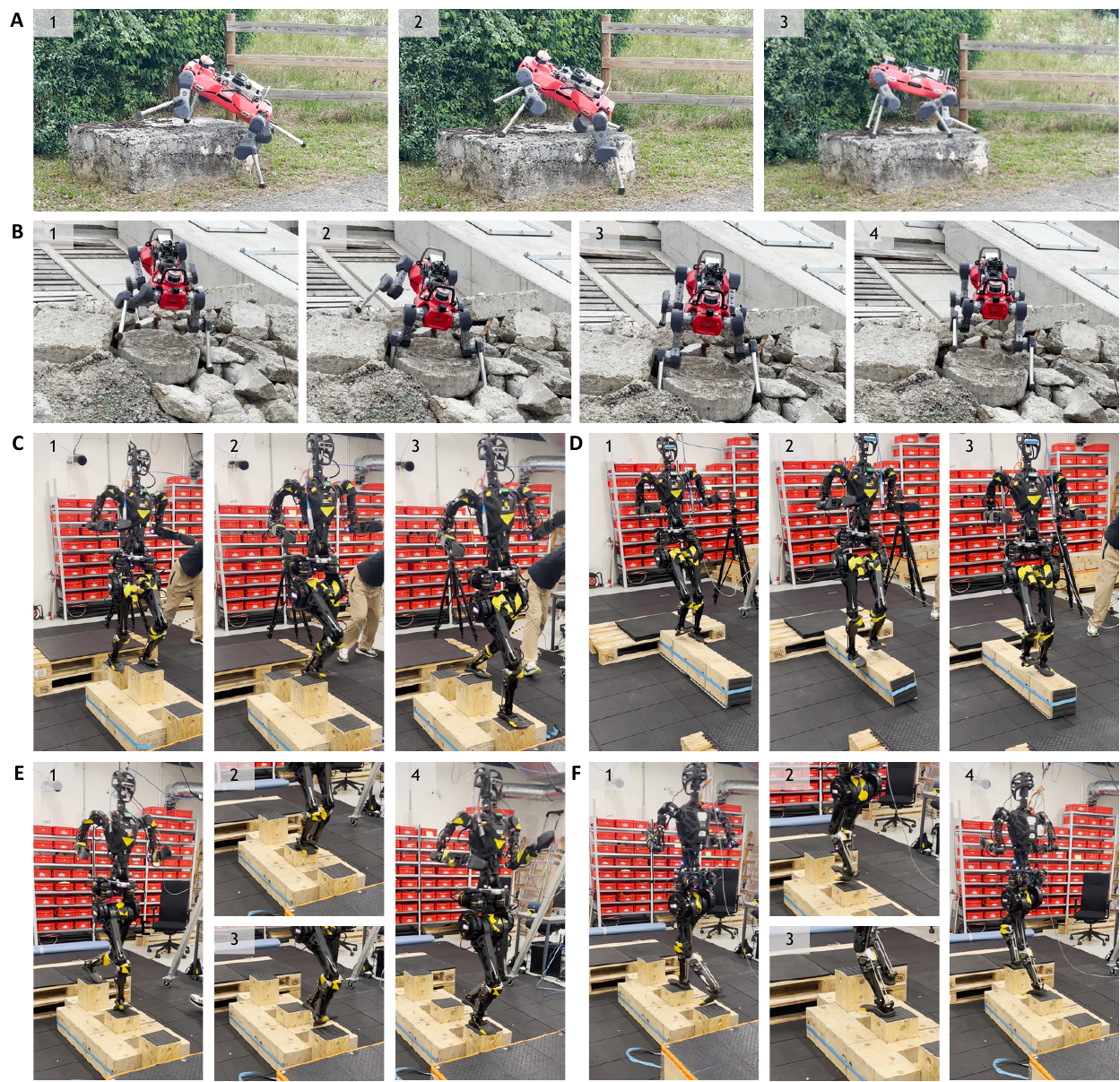}
\caption{\textbf{\highlight{Agility and Recovery Reflexes by Whole-Body Coordination.}} \textbf{(A)} ANYmal-D using its knee to climb up a large rock while rotating the torso. \textbf{(B)} ANYmal-D recovered from penetration of feet into the shaky debris due to slippage by knee support. \textbf{(C)} GR-1 traversing one row of 19-cm-wide uneven stepping stones with natural arm swing to aid the agile motions. \textbf{(D)} GR-1 stabilizing itself on a shaky, unfixed 19-cm-wide balance beam. \textbf{(E)} GR-1 encountered a slippage while traversing one row of uneven stepping stones and reacted with a fast step forward. \textbf{(F)} GR-1 encountered an inappropriate foothold due to left-biasing velocity commands while traversing a row of uneven stepping stones. With insufficient space for the left foot to land after the right foot's placement, GR-1 switched the foot contact in the air and successfully reached the subsequent stone with the right foot.}
\label{fig:emergent}
\end{figure*}

Our learned policies demonstrated advanced agility and robustness on the real robots, as shown in Figure~\ref{fig:emergent}. Learning whole-body motion control has aided ANYmal-D and GR-1 in actively using the knees (Figure~\ref{fig:emergent} A) and arms (Figure~\ref{fig:emergent} C), respectively, for enhanced agility. The amplitude and frequency of GR-1's arm swing not only follow the gait but also depend on the terrain, as can be observed by comparing Figure~\ref{fig:emergent} C to Figure~\ref{fig:emergent} D. The emergent recovery behaviors can save the robots from slippage (Figure~\ref{fig:emergent} B, E) or stabilize themselves on shaky supports (Figure~\ref{fig:emergent} D). Most remarkably, when the current footholds and the velocity commands made the next step hard to land (Figure~\ref{fig:emergent} F), we observed that GR-1 performed a single-leg switch hop on one stepping stone, lifting the previous contact foot into the air to successfully reach the subsequent stone. All of these reflex behaviors, which enhance the system stability, are difficult to obtain from model-based methods, as they typically rely on contact state machines and handcrafted heuristics~\cite{Jenelten2022, Ruben2022, chignoli2021humanoid, fahmi2022vital}.

\subsection{Versatile Velocity Tracking}
\begin{figure*}
\centering
\includegraphics[width=7.3in]{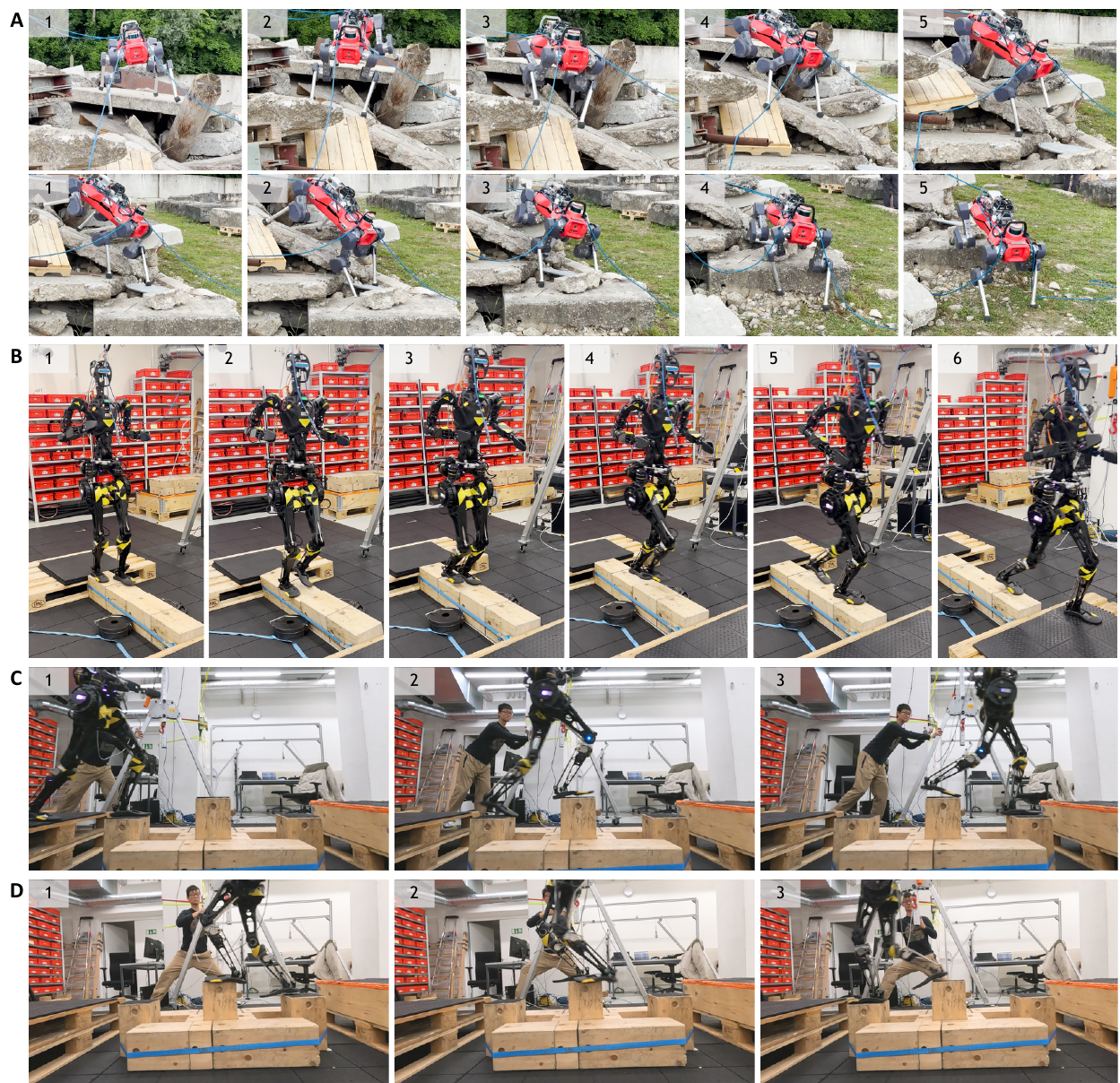}
\caption{\textbf{\highlight{Versatile Velocity Tracking.}} Our learned controllers demonstrated versatile velocity tracking capabilities on ANYmal-D and GR-1. \textbf{(A)} ANYmal-D maneuvering on the debris, overcoming sparse terrain with movable supports and showcasing omnidirectional versatility. \textbf{(B)} GR-1 accelerating on the shaky balance beam when the velocity command changed from 0.7 m/s to 1.5 m/s, taking longer strides. \textbf{(C)} With a 1.5 m/s forward velocity command, GR-1 had one step per stepping stone. \textbf{(D)} With a 0.7 m/s forward velocity command, GR-1 had two steps on each stepping stone.   }
\label{fig:versatile}
\end{figure*}

Our learned policies can track velocity commands in a versatile way, with detailed evaluations and benchmarks shown in Section~\ref{sec:evaluation}. This enables us to maneuver ANYmal-D on debris. In Figure~\ref{fig:versatile} A, the robot overcame sparse terrains with movable supports and showcased omnidirectional versatility. We can also command GR-1 with different velocities on challenging terrains, such as one row of uneven stepping stones or shaky balance beams, leading to different gait patterns and whole-body behaviors, as depicted in Figure~\ref{fig:versatile} B and C. 
Such versatility can expand the operational range of our robots in complex terrains and confined spaces.

\subsection{Simulation-based Evaluations}
\label{sec:evaluation}

\begin{figure*}
\centering
\includegraphics[width=7.3in]{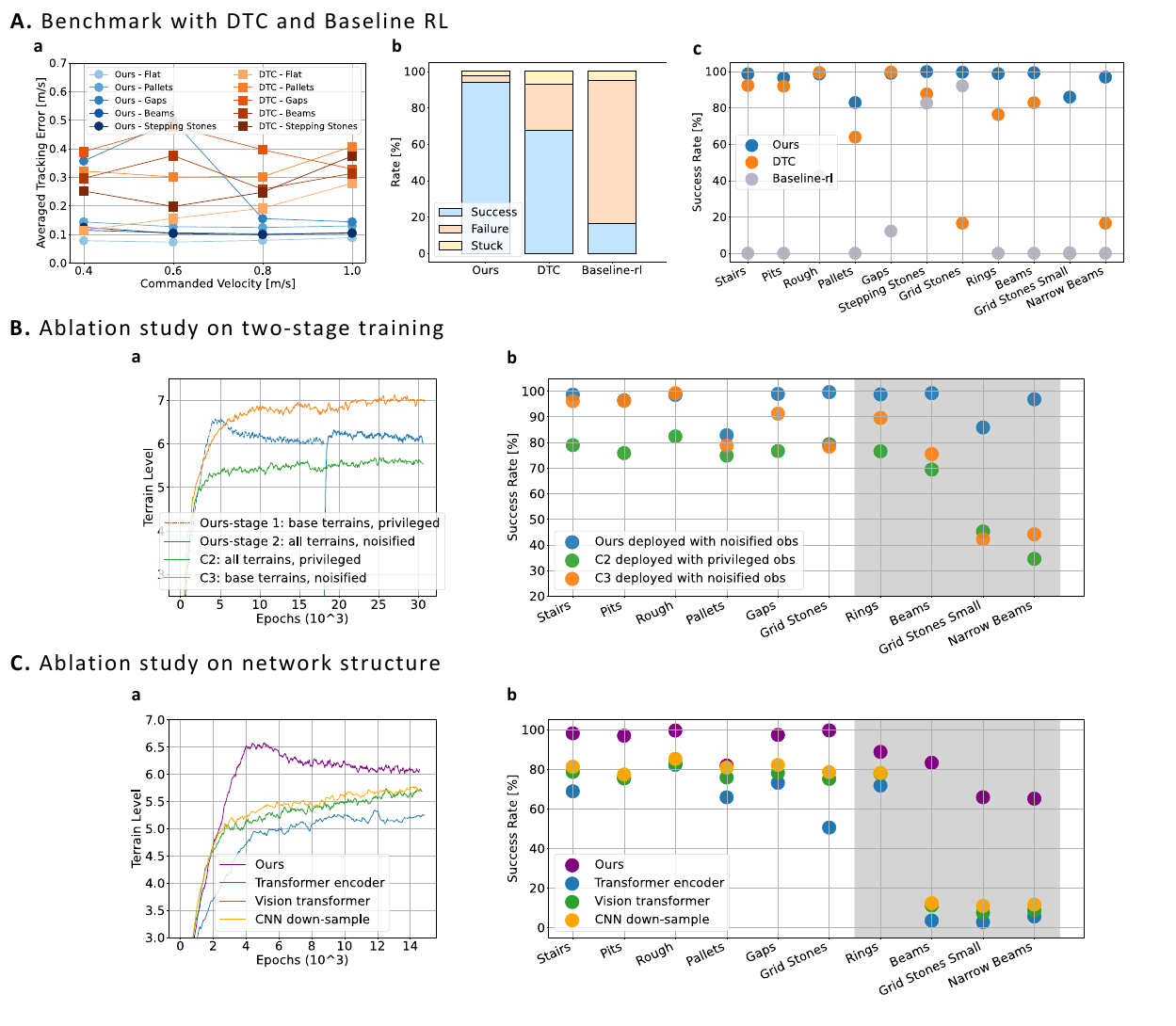}
\caption{\textbf{\highlight{Simulation-based Evaluations.}} We only evaluated the performance of our approach and benchmarked it with other methods on ANYmal-D. \textbf{(A) Benchmark with DTC and Baseline-rl.} \textbf{(a)} Our method shows overall lower velocity tracking errors for different forward velocity commands on the selected terrains. \textbf{(b)} Our method shows a substantially higher success rate, and lower stuck and failure rates (a trial 
 is counted as “successful” if the robot could walk out of the border of the terrain within a complete episode, “failed” if undesirable contacts happen, and “stuck” otherwise.) on a combination of all training terrains. \textbf{(c)} Our method demonstrates higher overall success rates on individual terrains. \textbf{(B) Ablation study on two-stage training.}  \textbf{(a)} Terrain level training curves for the proposed two-stage training (Ours), training from scratch on all terrains (base + fine-tuning terrains) with privileged observations (C2), and training from scratch on the base terrains with sensory drift and noise (C3). Ours shows the best convergent behavior. \textbf{(b)} Our method shows higher overall success rates on individual terrains, where the white background indicates the base terrains and the gray background fine-tuning terrains. \textbf{(C) Ablation study on network structure.} \textbf{(a)} Terrain level training curves for different methods on base terrains. Ours shows the best convergent behavior. \textbf{(b)} Our method shows higher overall success rates on individual terrains.}
\label{fig:evaluation}
\end{figure*}

\subsubsection{Benchmark with DTC and baseline RL controller}
In the first evaluation experiment, we evaluate the performance of three different controllers that \textbf{can dynamically navigate on sparse terrains} with an ANYmal-D -- the proposed method, DTC \cite{jenelten2023dtc}, and the baseline-rl~\cite{chong_iros}, which is a recent learning-based controller that demonstrates successful hardware experiments on stepping stones and beams. To do so, we utilized three comparing criteria similar to \cite{jenelten2023dtc} including \textbf{1)} the velocity tracking performances, \textbf{2)} success, failure, and stuck rates on terrains that the compared controllers are trained on, and \textbf{3)} the success rates on individual terrains. All controllers are deployed in the same simulated environment with observation noises and drifts sampled with the same random seed.

Since baseline-rl is designed under a goal-reaching setup, and there is currently no learning-based velocity-tracking controller tailored for sparse terrains that can parallel DTC's performance on hardware in the literature, we only compare the proposed method and DTC for velocity tracking, as shown in Figure \ref{fig:evaluation} A a. The agents with different controllers are deployed with constant forward velocity commands on selected sparse terrains traversable by both controllers. The tracking error is only computed for surviving agents. Our approach demonstrated substantially lower tracking errors except for gaps with small velocity commands, where the agents hesitate to proceed for both controllers. Noticeably, DTC exhibited high tracking errors with large velocity commands. This is mainly because DTC uses constant gait frequency, resulting in further away and thus harder-to-track footholds when velocity commands are higher. Our method, on the other hand, can adjust gait frequencies on different terrains with various velocity commands. Figure \ref{fig:evaluation} A b shows the success, failure, and stuck rates on a complete set of terrains that the three controllers were trained on, where our method shows 26.5\% and 77.3\% higher success rates compared to DTC and baseline RL, respectively. The reason for the low overall success rates of DTC and baseline RL can be explained in Figure \ref{fig:evaluation} A c. DTC has lower than 20\% success rates on grid stones (20cm $\times$ 20cm randomly placed square stones), grid stones small (12cm $\times$ 12cm randomly placed square stones), and narrow beams (15cm width), which is mainly because the high-level controller provides infeasible foothold guidance on these terrains since the support surface is smaller than the threshold that the model-based planner can accept as a steppable area. The baseline RL controller is overfitted to grid stones, explaining its low success rates on other terrains.

\subsubsection{Ablation study on two-stage training}
\label{sec:ablation_two_stage}
To justify the proposed two-stage training pipeline and the necessity of initializing the map encoding learning in stage 1 and introducing more terrains and uncertainties only in stage 2, we closely investigated and compared the training as well as deployment performance of three controllers. The first controller is trained with the proposed pipeline: first trained on base terrains with privileged observations (ground truth) and then fine-tuned on all terrains (base terrains and fine-tuning terrains) with sensory drifts and noises. The second controller (C2) is trained on all terrains from scratch with privileged observations, while the third controller (C3) is trained on base terrains from scratch with sensory drifts and noises. We use terrain level~\cite{Rudin2021} to compare their training performance. The terrains have 10 difficulty levels, indexed from 0 to 9. All robots are randomly assigned a terrain type and a level. The robot gets upgraded to the next level if it walks out of the borders of its assigned terrain, and downgraded to a lower level otherwise. Robots solving the highest level are then reset to a randomly selected level, which averages to level 6 if all robots have solved the most difficult terrain. Perfect training will show a terrain level curve that first overshoots over level 6 (robots attempting to upgrade to higher terrain levels) and then converges to level 6 (based on the stationary distribution when all robots solve their terrains). Figure \ref{fig:evaluation} B a indicated that our proposed method resulted in a convergent terrain level that shows most of the robots solve the most difficult level. However, C2 could not reach the same terrain level, meaning the agents failed to upgrade to higher terrain levels. C3 could not converge back to 6, indicating that the agents could not solve the most difficult levels. We then deployed the controllers and compared the success rates on different terrains, as depicted in Figure \ref{fig:evaluation} B b. Our method shows substantially higher success rates than C2 and C3 on almost all terrains. C2 shows the worst performance even with perfect perception during testing. While C3 shows satisfactory success rates on stairs, pits, and rough, it performs worse on sparse terrains (grid stones, beams, etc). 

\subsubsection{Ablation study on network structure}
Our attention-based map encoding consists of two levels: 1) a low-level CNN that embeds local terrain features, and 2) an MHA module that queries point-wise local features and combines them with proprioceptive observations. To show the efficacy of using an MHA module the way we propose, we compare our method to a transformer encoder similar to~\cite{yang2022learning}. To show the necessity of point-wise attention, we down-sampled the map inputs with another CNN instead of only extracting local features without down-sampling. As another comparison, we used a vision transformer encoder~\cite{vit} to process the map scans. The detailed structures can be found in the supplementary methods section "Ablation Study Details - network structures". Similar to Section~\ref{sec:ablation_two_stage}, we compared the training and deployment performance of the above 3 network structures with our method through terrain level and success rates, as shown in Figure \ref{fig:evaluation} C a and b, respectively. From the comparison, our method demonstrates higher convergent terrain levels during training and success rates during deployment. More notably, our method shows substantially higher success rates on unseen terrains, indicating better generalization than the other potential network structures.

\subsection{Interpretable Attention-based Map Encoding}
\begin{figure*}
\centering
\includegraphics[width=7.0in]{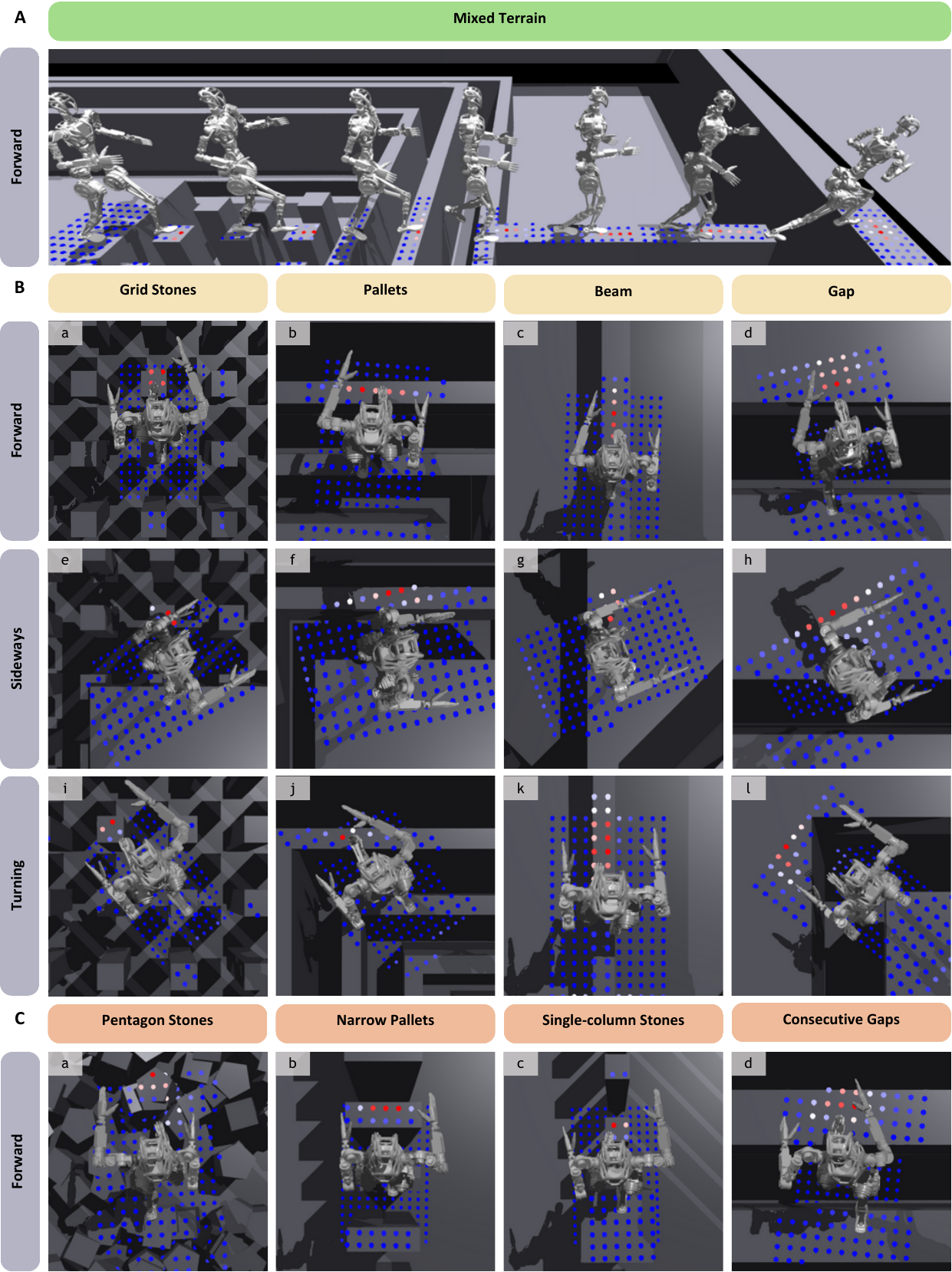}
\caption{\textbf{\highlight{Attention Weights Visualization.}} We visualized the height scans with associated attention weights on each scan, where colors with higher intensity in red indicate higher attention. \textbf{(A)} Attention weights on a mixture of terrains for a stage 2 controller. \textbf{(B)} Attention weights on individual base terrains for a stage 1 controller with forward, sideways, and turning velocity commands. \textbf{(C)} Attention weights on individual fine-tuning terrains for a stage 1 controller with forward velocity commands.}
\label{fig:attn_weights}
\end{figure*}

To further elucidate the interpretability provided by the proposed attention-based map encoding, we provide a detailed visualization of the attention weights from the MHA module across different types of terrains. The MHA module pays more attention to steppable areas based on the proprioceptive information of the robots and allocates higher attention weights to corresponding map scans. These visualizations highlight how the model prioritizes specific regions in the environment, guiding the controller to navigate complex discontinuous terrains. Figure \ref{fig:attn_weights} A showcases the attention weights of the fine-tuned controller from stage 2 on a mixed terrain, which combines elements from various terrain types. The attention weights are concentrated around the next steppable region, indicating that the MHA module acts as guidance to direct the controller toward feasible footholds ensuring stability. Figure \ref{fig:attn_weights} B depicts the attention weights when the robots are commanded to move in different directions on various terrains encountered during stage 1 training, including grid stones (B 1 forward, B 5 sideways, B 9 turning), pallets (B 2 forward, B 6 sideways, B 10 turning), single beams (B 3 forward, B 7 sideways, B 11 turning), and gaps (B 4 forward, B 8 sideways, B 12 turning). These terrains, characterized by their varying structural features, challenge the model to distribute attention effectively to critical regions to support footholds based on the robots' kinematics and dynamics. Noticeably, the paid attention can refuse to follow infeasible velocity commands. For example, the agent is commanded to turn but the attention still stays on the beam rather than on the ground to follow the command, as demonstrated in Figure \ref{fig:attn_weights} B k. Additionally, Figure \ref{fig:attn_weights} C illustrates how the stage 1 controller generalizes to unseen terrains, such as pentagon stones (C 1), narrow pallets (C 2), single-column stones (C 3), and consecutive gaps (C 4), highlighting the generalization of the attention mechanism in adapting to novel environments. These visualizations reflect the model’s ability to dynamically adjust its focus based on the robot's proprioceptive information and command directions, promoting efficient and safe locomotion across diverse and unpredictable terrains. The consistent pattern of attention allocation supports our claim that the MHA module enhances both interpretability and generalization capabilities in complex locomotion tasks.\\

\section*{Discussion}



In this work, we achieved generalized, agile, and robust legged locomotion using end-to-end reinforcement learning, mapping proprioceptive and exteroceptive observations directly to joint-level actions with a neural network. The key to achieving this is an attention-based map encoding module which processes the map based on proprioception and provides a representation that learns to focus on potential future footholds. The subsequent policy then translates this generalized representation into whole-body motions, enabling precise movements on sparse terrains. Additionally, we developed a two-stage training pipeline to further enhance the generalization capability of the controller. The resulting controllers enabled both quadrupedal and humanoid robots to traverse diverse terrains and demonstrate highly interpretable neural encoding of terrain perception.

We enabled generalized legged locomotion while preserving the robustness of learning-based controllers. Previously, such generalization only appeared on quadruped locomotion controllers with model-based planning~\cite{jenelten2023dtc, Ruben2022}. Hence, on top of the SOTA performance on quadrupedal and humanoid robots, this work demonstrates the possibility of achieving comparable generalization using DRL while overcoming the limits of model-based approaches when dealing with uncertainties and model errors.

Notably, our policy network design mirrors the modular functions in model-based methods: the map encoding module implicitly selects future footholds like contact planners by attention, and the subsequent policy acts like a whole-body controller tracking the planned contacts. Yet, by integrating the entire controller end-to-end, we can leverage data-driven learning to tackle the issues of model-based methods: computational burden, model mismatch, and violation of assumptions. We can also tune the controller as a whole rather than tuning separate modules, reducing system complexity.

This work also has limitations. First, training the policies can take several days, and parameter tuning becomes inefficient due to the training costs. Second, we use the 2.5D height map representation which may be inapplicable to certain scenarios, such as confined spaces~\cite{10610271_miki_3d}. Third, we focus on locomotion in this work, while the arms need to be used for manipulation as well, and we have not studied how to balance the needs for both locomotion and manipulation with the arms.

In the future, we will explore how we can improve training efficiency and develop effective 3D representations that can generalize to a broader range of scenarios. We also expect that the attention mechanism can be extended to loco-manipulation tasks, including opening doors, moving obstacles, climbing with the help of hands, etc.

\section*{Materials and Methods}

\begin{figure*}
\centering
\includegraphics[width=7.3in]{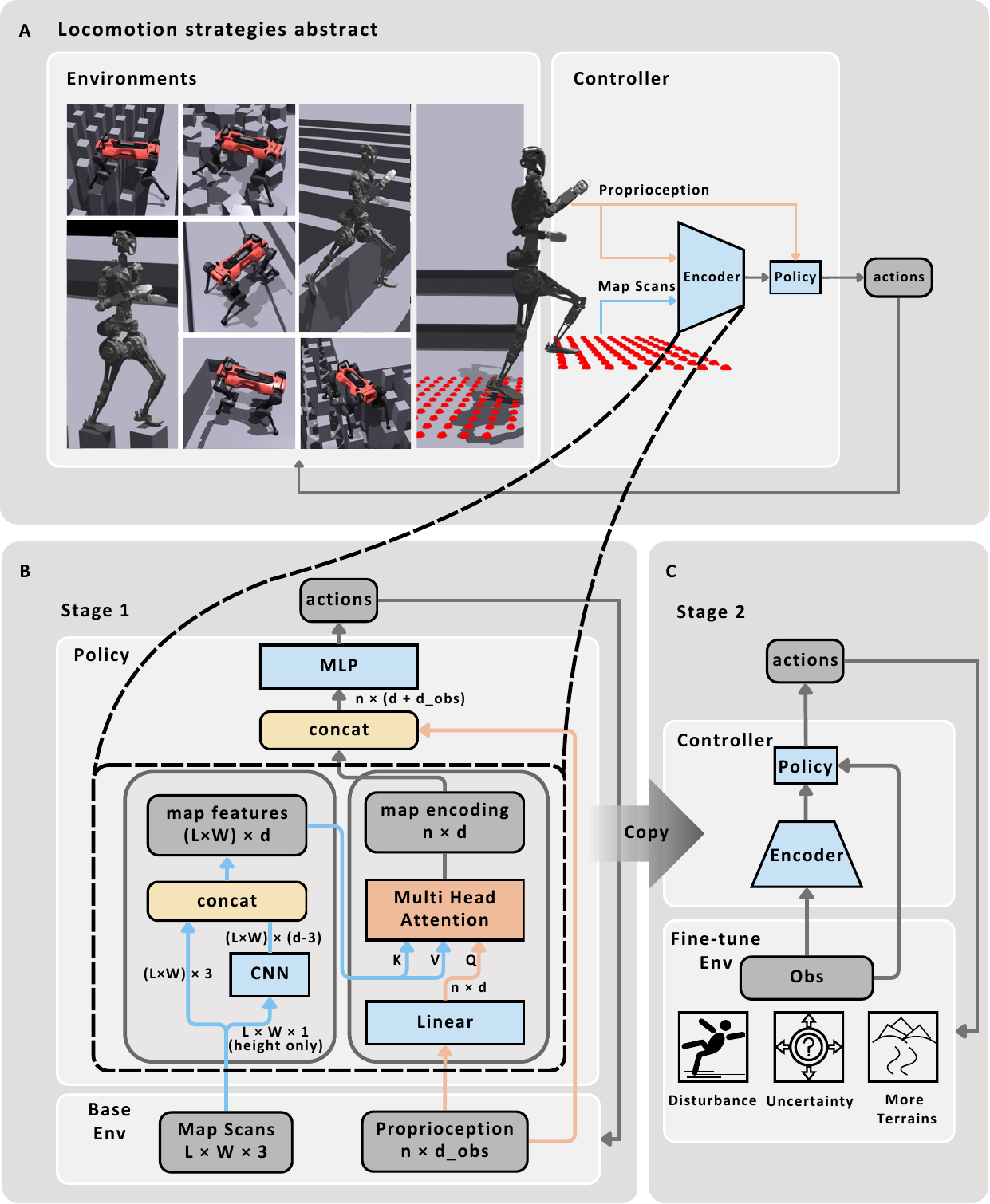}
\caption{\textbf{\highlight{Proposed Control Pipeline.}} \textbf{(A)} A locomotion strategies abstract. \textbf{(B)} Stage 1 training.\textbf{(C)} Stage 2 fine-tuning with disturbance, uncertainty, and extra harder terrains. }
\label{fig:intro}
\end{figure*}

\subsection*{\highlight{Motivation}}
A general abstract model can be derived to describe the commonalities among model-based, learning-based, and hybrid methods as pictured in Figure \ref{fig:intro} A. In summary, an encoder observes proprioceptive and exteroceptive information and outputs a latent representation that is then fed into subsequent policy modules to generate actions. 
 
For model-based controllers, the encoder can be interpreted as a high-level planner that predicts the base trajectories, joint angles, and footholds within its prediction horizon. The subsequent policy module can be interpreted as a whole-body controller that generates corresponding actions to track the solutions from the planner. Similarly, for learning-based controllers, Rudin et al.~\cite{Rudin2021} use a single MLP to embed the observations and generate actions. Miki et al.~\cite{Miki2022} introduced a belief encoder to reconstruct exteroception when sensor inputs are degraded, subsequently passing the reconstructed representation to the subsequent module. For hybrid methods, some works\cite{Tsounis2020, RLOC} utilize a network as the encoder to generate footholds, with a model-based whole-body controller serving as the subsequent module to track these footholds. Jenelten et al.\cite{jenelten2023dtc} leverage TAMOLS as the encoder to generate base trajectories and footholds and use an MLP policy to track TAMOLS solutions. Despite the variety in these locomotion strategies, they share intrinsic similarities in their control pipelines. 

Building on this general abstract framework, we can analyze the factors contributing to differences in robustness, precision, and generalizability among the control strategies mentioned above. Model-based controllers have exhibited great precision, with model-based planners serving as the observation encoder. Model-based planners leverage trajectory optimization to predict footholds and base motions from map scans and robot states while adhering to the kinematic and dynamic constraints. However, they tend to be less robust due to their inability to handle model mismatches and uncertainties effectively. In contrast, learning-based methods demonstrate greater robustness, as their encoders employ neural networks, which are better at managing uncertainties when trained with proper domain randomization. Nevertheless, they are generally less precise, as it is difficult to impose hard constraints on neural networks, and less generalizable to new terrains, given their susceptibility to overfitting. With these considerations, we focus on the encoder as a key component in developing a control strategy that balances precision, robustness, and generalizability across terrains. In this work, we introduce an attention-based neural locomotion controller architecture that employs a multi-head attention module to encode exteroception conditioned on proprioception, and thus is capable of predicting precise foot placements based on current robot states, as illustrated in Figure \ref{fig:intro} B.

\subsection*{Multi-Head Attention-Based Map Encoding}
Multi-head attention (MHA) \cite{NIPS2017_3f5ee243} is a neural representation mechanism that has revolutionized deep learning. At its core, MHA enables the model to focus on the important parts of the inputs, and enriches the representation by generating multiple parallel attention outputs, colloquially known as "heads". MHA thereby enhances interpretability by identifying important inputs and improves generalization by mitigating the effects of variations in less important inputs. These features align with our requirements for interpretability and generalization in locomotion controllers that can work across diverse terrains.

Technically, the inputs of MHA in our work include a query and a collection of key-value pairs. The output is determined by calculating a weighted sum of the values, with each value’s weight being determined by the compatibility of the query with the corresponding key. This process can be interpreted as paying more attention to the values, for which the corresponding keys are the most relevant to the query inputs. This interpretation can be naturally transferred to our case, where we expect the MHA module to focus more on the optimal steppable areas, informed by current proprioceptive data and commands. To this end, we use the proprioception embedding as the query and point-wise local features as the key-value pairs to get a map encoding conditioned on the proprioception. This structure is more expressive, or put another way, more suitable to model intricate relationships between the proprioceptive and exteroceptive information than MLPs because the attention weights paid to the map scans are state-dependent. The outputs of MHA are then translated to joint-level actions by the subsequent policy module.

As demonstrated in Figure \ref{fig:intro} B, the map scans (L $\times$ W $\times$ 3, L points long, W points wide, 3-d coordinates in the robot frame for each point) make the exteroceptive observations. The z-values of the points are first processed by a convolutional neural network (CNN) which consists of two layers with zero padding to keep the original dimensionality, and a kernel size of 5 to extract the local features for each point. The first layer has 16 hidden units and the second has $d - 3$ hidden units, where $d$ is the dimension of the MHA module (64 in our case). Then, we concatenate the output of CNN (L $\times$ W $\times$ ($d - 3$)) with the 3-d coordinates to get point-wise local features of shape LW $\times$ $d$. These local features extract the neighborhood features around each map point, enabling the subsequent MHA module to pay point-wise attention. 
 Meanwhile, the proprioception ($1 \times d_{obs}$) goes first through a linear layer that outputs a proprioception embedding ($1 \times d$). Then, with the proprioception embedding as the query (Q, with $n=1$ as MHA's query length) and the local features as the keys and values (K and V), the MHA module outputs the map encoding, with each head processing d/h dimensions of the inputs, where h is the number of heads.
 

\subsection*{Two-stage training pipeline}
To learn a robust and generalizable map encoding, we designed a two-stage training pipeline that progressively refines the controller’s capabilities. In the first stage, the controller is trained on base terrains with perfect perception. This stage warms up the map encoding learning and allows the controller to acquire locomotion skills using ground truth sensing. 

In the second stage, we introduce more complex terrains with disturbances and uncertainties. These terrains simulate more realistic environments where perception may be imperfect, shaping the learned motions to be more adaptable and resilient. By exposing the robot to a broader range of terrains with added disturbances, this stage improves the controller’s generalization across diverse terrains and also enhances its robustness against real-world uncertainties, which ensures that the final learned map encoding can operate effectively in unseen environments.

\subsection*{Observation Space}
The policy network observes proprioception information and map scans in the robot-centric base frame (torso link for ANYmal-D and pelvis link for GR-1 are considered as the base), including the base linear velocity $\boldsymbol{v}_{b}$, angular velocity $\boldsymbol{\omega}_{b}$, gravity vector $\mathbf{g}_{b}$, joint positions $\mathbf{q}_j$ and velocities $\mathbf{\dot{q}}_j$, previous actions $\mathbf{a}_{t-1}$ inferred by the control policy, and a vector map scans surrounding the robot's base (a total of L $\times$ W $\times$ 3 points, where L and W are length and width, respectively). The critic network observes the same information but without noise. The symbols are declared in Table ~\ref{tab:symbol}.\\

\begin{table}[h]
\centering
\caption{\textbf{Nomenclature}}
\label{tab:symbol}
\begin{tabular}{cl}
\hline

$\mathbf{v}_{i}^{*}$ & commanded base linear velocity of rigid body i         \\
$\mathbf{v}_{i}$     &  linear velocity of rigid body i\\
$\boldsymbol{\omega}_{i}^*$ &  commanded base angular velocity of rigid body i        \\
$\boldsymbol{\omega}_{i}$ &  angular velocity of rigid body i         \\
$\mathbf{a}_j$         &  desired $j$-th joint position inferred by the controller  \\
$\mathbf{q}_{0, j}$ &   default $j$-th joint position       \\
$\mathbf{q}_j$ &   joint positions      \\
$\mathbf{\dot{q}}_j$ &   joint velocities       \\
$\mathbf{\ddot{q}}_j$ &   joint accelerations   \\
$\boldsymbol{\tau}_j$ &   joint torques        \\
$\mathbf{q}_{lim, j}$ &  joint position limits \\
$\mathbf{\dot{q}}_{lim, j}$ &  joint velocity limits \\
$\boldsymbol{\tau}_{lim, j}$ &  torque limits \\
$n_{termination}$   &   number of terminations        \\
$n_{collision, i}$      &   number of collisions of rigid body i          \\
$n_{zero\_contact}$   &   number of zero contact        \\
$\mathbf{F}_f$                &  net contact force on foot f\\ 
$c_f$                &    contact state of foot f  \\
$\mathbf{v}_f$                &    velocity of foot f  \\ 
$\mathbf{g}_{i}$ & gravity vector of rigid body i         \\
$i_{*}$                &    index of rigid body named * \\\hline
\end{tabular}
\end{table}

\subsection*{Reward Functions}
The reward is a weighted sum of 14 terms for ANYmal-D and 16 terms for GR-1, detailed in Table~\ref{tab:rew_func}. We divided our reward settings into three categories, including task, regulation, and style rewards. The task rewards consist of tracking commanded linear and angular velocity while avoiding failures on terrains, where $n_{collision}$ is defined as the number of collisions between specific body parts with the terrain, and $n_{termination}$ is defined as the number of early terminations. We terminate the episode while there is a collision between the torso and the terrain, or bad orientation of the torso. The regulation rewards are designed to smooth the actions and avoid drastic motions, over-torque, and over-extensions. To generate more natural movements, we also introduced style rewards, where we penalize unwanted velocities, stomping, foot slippage, jumping gaits, and tilted body parts. To confine the movements of arms for GR-1, we also introduced a reward to penalize deviation from default joint positions above a specific threshold for arm joints. After we obtained a baseline controller using the reward functions above, we fine-tuned the controller in stage 2 to improve the sim-to-real transfer performance while standing, where we penalize joint motions while the robots are in the stance phase, as shown in orange in Table \ref{tab:rew_func}.

\begin{table*}[h]
\centering
\caption{\textbf{Definition of reward terms for ANYmal-D and GR-1}. The x-axis points forward and the z-axis points downward. The body, joint, and feet indices can be found in Appendix. The rewards marked in orange are only activated during stage 2 fine-tuning.}
\label{tab:rew_func}
\begin{tabular}{|crc|cc|cc|}
\hline
\multicolumn{2}{|c|}{}                                                                        &                                                                   & \multicolumn{2}{c|}{ANYmal-D}                                                             & \multicolumn{2}{c|}{GR-1}                                                                                                                                       \\ \cline{4-7} 
\multicolumn{2}{|c|}{\multirow{-2}{*}{Reward Terms}}                                          & \multirow{-2}{*}{functions}                                       & \multicolumn{1}{c|}{weights}                     & indices                                & \multicolumn{1}{c|}{weights}                     & indices                                                                                                      \\ \hline
\multicolumn{1}{|c|}{}                              & linear velocity tracking                & $\exp(-\lVert \mathbf{v}_{xy, i}^{*} - \mathbf{v}_{xy, i}\rVert^2)$                 & \multicolumn{1}{c|}{5.0}                         & $i = {[}i_{torso}{]}$               & \multicolumn{1}{c|}{5.0}                         & $i = {[}i_{torso}{]}$                                                                                       \\ \cline{4-7} 
\multicolumn{1}{|c|}{}                              & angular velocity tracking               & $\exp(-\lVert \boldsymbol{\omega}_{z, i}^{*} - \boldsymbol{\omega}_{z, i} \rVert^2)$ & \multicolumn{1}{c|}{3.0}                         & $i = {[}i_{torso}{]}$                 & \multicolumn{1}{c|}{3.0}                         & $i = {[}i_{torso}{]}$                                                                                       \\ \cline{4-7} 
\multicolumn{1}{|c|}{}                              & termination penalty                     & $-n_{termination}$                                                & \multicolumn{1}{c|}{200}                         & N/A                                    & \multicolumn{1}{c|}{200}                         & N/A                                                                                                          \\ \cline{4-7} 
\multicolumn{1}{|c|}{\multirow{-4}{*}{Tasks}}       & collision penalty                       & $-n_{collision, i}$                                               & \multicolumn{1}{c|}{1}                           & $i = {[}i_{shank}{]}$                & \multicolumn{1}{c|}{\cellcolor[HTML]{9B9B9B}}    & \cellcolor[HTML]{9B9B9B}                                                                                     \\ \hline
\multicolumn{1}{|c|}{}                              & action rate                             & $-||\mathbf{a_j}_t - \mathbf{a_j}_{t-1}||^2$                      & \multicolumn{1}{c|}{5.0e-3}                      & j = {[}0:12{]}                         & \multicolumn{1}{c|}{5.0e-3}                      & j = {[}0:23{]}                                                                                               \\ \cline{4-7} 
\multicolumn{1}{|c|}{}                              & joint acceleration penalty              & $-||\mathbf{\ddot{q}}_j||^2$                                      & \multicolumn{1}{c|}{2.5e-7}                      & j = {[}0:12{]}                         & \multicolumn{1}{c|}{1e-6}                        & j = {[}15:18, 19:22{]}                                                                                       \\ \cline{4-7} 
\multicolumn{1}{|c|}{}                              &                                         &                                                                   & \multicolumn{1}{c|}{}                            &                                        & \multicolumn{1}{c|}{1e-4}                        & j = {[}15, 19{]}                                                                                             \\
\multicolumn{1}{|c|}{}                              &                                         &                                                                   & \multicolumn{1}{c|}{}                            &                                        & \multicolumn{1}{c|}{5e-5}                        & j = {[}3,9{]}                                                                                                \\
\multicolumn{1}{|c|}{}                              & \multirow{-3}{*}{joint torques penalty} & \multirow{-3}{*}{\&   $-||\boldsymbol{\tau}_j||^2$}               & \multicolumn{1}{c|}{\multirow{-3}{*}{2.0e-5}}    & \multirow{-3}{*}{j = {[}0:12{]}}       & \multicolumn{1}{c|}{5e-5}                        & j = {[}2,8{]}                                                                                                \\ \cline{4-7} 
\multicolumn{1}{|c|}{}                              & joint position limits                   & $- max (|q_j| - 0.9q_{lim, j}, 0)$                   & \multicolumn{1}{c|}{1.0} & j = {[}0:12{]} & \multicolumn{1}{c|}{10}                          & j = {[}0:23{]}                                                                                               \\ \cline{4-7} 
\multicolumn{1}{|c|}{}                              & joint velocity limits                   & $- max (|\dot{q}_j| - 0.9\dot{q}_{lim, j}, 0)$    & \multicolumn{1}{c|}{1.0}                         & j = {[}0:12{]}                         & \multicolumn{1}{c|}{0.1} & j = {[}0:23{]}                                                                       \\ \cline{4-7} 
\multicolumn{1}{|c|}{\multirow{-8}{*}{Regulations}} & joint torque limits                     & $- max (|\tau_j| - 0.8\tau_{lim, j}, 0)$          & \multicolumn{1}{c|}{0.2}                         & j = {[}0:12{]}                         & \multicolumn{1}{c|}{2e-3}                        & j = {[}0:23{]}                                                                                               \\ \hline
\multicolumn{1}{|c|}{}                              & linear velocity penalty                 & $-\mathbf{v}_{z, i}^2$                                             & \multicolumn{1}{c|}{1.0}                         & $i = {[}i_{torso}{]}$                 & \multicolumn{1}{c|}{\cellcolor[HTML]{9B9B9B}}    & \cellcolor[HTML]{9B9B9B}                                                                                     \\ \cline{4-7} 
\multicolumn{1}{|c|}{}                              & angular velocity penalty                & $-||\boldsymbol{\omega}_{xy, i}||^2$                               & \multicolumn{1}{c|}{5.0e-2}                      & $i = {[}i_{torso}{]}$                 & \multicolumn{1}{c|}{5.0e-2}                      & $i = {[}i_{torso}{]}$                                                                                       \\ \cline{4-7} 
\multicolumn{1}{|c|}{}                              & contact forces penalty                  & $-max(||\mathbf{F}_f|| - 700, 0)$                  & \multicolumn{1}{c|}{2.5e-5}                      & f = {[}0:4{]}                          & \multicolumn{1}{c|}{\cellcolor[HTML]{9B9B9B}}    & \cellcolor[HTML]{9B9B9B}                                                                                     \\ \cline{4-7} 
\multicolumn{1}{|c|}{}                              & foot slippage penalty                   & $-c_f * ||\mathbf{v}_f||$                          & \multicolumn{1}{c|}{0.5} & f = {[}0:4{]}                          & \multicolumn{1}{c|}{1.0}                         & f = {[}0, 1{]}                                                                                               \\ \cline{4-7} 
\multicolumn{1}{|c|}{}                              & joint deviation penalty                 & $ max (||q_j - q_{0, j}||^2 - 0.25, 0.0)$                              & \multicolumn{1}{c|}{\cellcolor[HTML]{9B9B9B}}    & \cellcolor[HTML]{9B9B9B}               & \multicolumn{1}{c|}{0.5}                         & j = {[}15:23{]}                                                                                              \\ \cline{4-7} 
\multicolumn{1}{|c|}{}                              & no fly                                  & $-n_{zero\_contact}$                                          & \multicolumn{1}{c|}{\cellcolor[HTML]{9B9B9B}}    & \cellcolor[HTML]{9B9B9B}               & \multicolumn{1}{c|}{5.0}                         & f = {[}0, 1{]}                                                                                               \\ \cline{4-7} 
\multicolumn{1}{|c|}{}                              & straight body                           & $-||g_i||^2$                                                      & \multicolumn{1}{c|}{\cellcolor[HTML]{9B9B9B}}    & \cellcolor[HTML]{9B9B9B}               & \multicolumn{1}{c|}{3.0}                         & \begin{tabular}[c]{@{}c@{}}$i = {[}i_{torso}$, \\         $i_{pelvis}$, \\     $i_{feet}{]}$\end{tabular} \\ \cline{4-7} 
\multicolumn{1}{|c|}{}                              & standing joint positions penalty        & $-||\mathbf{q}_{j}^{*} - \mathbf{q}_{j}||$                        & \multicolumn{1}{c|}{\cellcolor[HTML]{FFCE93}0.1} & \cellcolor[HTML]{FFCE93}j = {[}0:12{]} & \multicolumn{1}{c|}{\cellcolor[HTML]{9B9B9B}}    & \cellcolor[HTML]{9B9B9B}                                                                                     \\ \cline{4-7} 
\multicolumn{1}{|c|}{\multirow{-9}{*}{Styles}}      & standing joint velocity penalty         & $-||\mathbf{\dot{q}}_{j}^{*} - \mathbf{\dot{q}}_{j}||$            & \multicolumn{1}{c|}{\cellcolor[HTML]{FFCE93}0.5} & \cellcolor[HTML]{FFCE93}j = {[}0:12{]} & \multicolumn{1}{c|}{\cellcolor[HTML]{FFCE93}0.2} & \cellcolor[HTML]{FFCE93}j = {[}0:23{]}                                                                       \\ \hline
\end{tabular}
\end{table*}

\subsection*{Training Environment}
We utilize a custom version of Proximal Policy Optimization (PPO)~\cite{ppo} for training, and a two-stage training pipeline, where we first train a controller with privileged observations for both the actor and the critic and then fine-tune the controller with noises and disturbances for the actor and privileged information for the critic. The actor and the critic use the same network structure as shown in Figure~\ref{fig:intro} B, where they share the same encoder module but use different subsequent MLPs. The actor MLP maps the MHA output to joint actions while the critic MLP maps to a value. The hyperparameters are detailed in the supplementary methods (section "Training Details - PPO parameters"). 

\subsubsection*{Domain Randomization}
We introduce noise to observations that are not privileged. At each simulation step, a customized noise is sampled from a uniform distribution and added to each observation term, except for the previous actions and velocity commands. The map scans also have random drifts sampled from a normal distribution for each terrain at the beginning of training. The perturbations are mainly designed to improve the robustness against sensor drifts during deployment. We also introduced artificial pushes by resetting the twist of the robots in simulation. To improve the controller's robustness against payload and friction variations, we also randomized the torso mass and the friction coefficient of each contact foot. 

\subsubsection*{Terrains and Curriculum}
We use different terrain settings for each training stage, detailed in the supplementary methods (section "Training Details - Terrains"). For each training stage, we use a curriculum introduced in ~\cite{Rudin2021}. At the start of the training, all robots are randomly assigned a terrain type and a difficulty level (10 in total). During training, the robots that managed to walk out of their assigned terrains get upgraded to the next level and downgraded to a lower level otherwise. The difficulty level of each terrain is tuned heuristically such that the supporting surface is big enough for the map scan resolution and the difficulty is not over the maximum robot capability. For example, for a 10cm map resolution, the stepping stones should be larger than 10cm; the gap width should not exceed the length of the quadruped robot/the maximum possible feet distance of the humanoid robot while assuming a walking gait (at least one foot on the ground).  

\subsection*{Training}
The network parameters are the same for ANYmal-D and Fourier GR-1, except that the dimension of map scans for GR-1 (17 $\times$ 11) is smaller than that of ANYmal-D (26 $\times$ 16). We used $d=64$ for the MHA dimension, $n=1$ for the target sequence length, and $h = 16$ for the number of heads. We then trained the ANYmal controller through massive parallelization with 4096 robots for 18000 epochs in stage 1 and 3600 epochs in stage 2. With 24 seconds of training time per epoch, the total training time is 6 days on an Nvidia Tesla A100-40GB GPU, which is a reduction of roughly $60\%$ of training time compared to DTC. For the GR-1 controller, we trained for 15000 epochs in stage 1 and 3200 epochs in stage 2, with 14 seconds per epoch, resulting in a total training time of 3.5 days on an Nvidia RTX 4090 GPU.

\subsection*{Deployment}
We used ANYmal-D and GR-1 for our experiments. For ANYmal-D, the control policy inference is done on a single Intel core-i7 8850H CPU, and the elevation mapping~\cite{elevationmapping_cupy} runs on an onboard Nvidia Jetson. For GR-1, the policy inference is done on the Intel core-i7 13700h CPU, and the map scans are sampled through ray-casting on a pre-designed terrain mesh given the robot's pose captured by the Qualysis Motion Capture system~\cite{qtm}.



\bibliography{scibib}

\section*{Acknowledgments}
We thank Nikita Rudin and Vladlen Koltun for their helpful discussion. 
\textbf{Funding:}
This work was funded in part by the NCCR Automation, EU Project 101121321 and 852044, the ETH Mobility Initiative, Fourier Intelligence, and Apple Inc. Any views, opinions, findings, and conclusions or recommendations expressed in this material are those of the authors and should not be interpreted as reflecting the views, policies or position, either expressed or implied, of Apple Inc. 
\textbf{Author contributions:} 
 J.~H. and C.~Z. formulated the main ideas, trained the controllers, and conducted most of the experiments. F.~J. and R.~G. helped with experiments and provided insights into system development. M.~B. and M.~H. acquired the funding for the project. All authors helped to write, improve, and refine the paper.  
\textbf{Competing interests:}
The authors declare that they have a patent application pending. 
\textbf{Data and materials availability:} Data to reproduce our plots are available \cite{data}.

\clearpage

\section*{\highlight{Supplementary Methods}}
\setcounter{figure}{0}
\renewcommand\thefigure{S\arabic{figure}}
\setcounter{table}{0}
\renewcommand\thetable{S\arabic{table}}


\subsection*{Training Details}

\subsubsection*{Terrains}
ANYmal-D:
\begin{itemize}
    \item \textbf{base terrains} (terrains for stage 1 training)
    \begin{itemize}
        \item stairs
        \item pits - stages with a certain height
        \item rough - ground with $\pm$ 8cm height noise
        \item pallets - horizontally placed pallets with random distances and heights
        \item gaps
        \item grid stones - randomly placed stepping stones with random height differences
    \end{itemize}
    \item \textbf{fine-tuning terrains} - terrains added for stage 2 fine-tuning
    \begin{itemize}
        \item pentagon stones - randomly placed stones with pentagon surfaces.
        \item rough hills - inclined rough ground
        \item rings - ring-shaped steps
        \item beams - beams radiating outward from the center
        \item grid stones small - grid stones with a minimum of 12cm width
        \item narrow beams - beams with a minimum of 15cm width
    \end{itemize}
\end{itemize} 

\noindent GR-1:
\begin{itemize}
    \item \textbf{base terrains} (terrains for stage 1 training)
    \begin{itemize}
        \item stairs
        \item pits - stages with a certain height
        \item rough - ground with $\pm$ 8cm height noise
        \item pallets - horizontally placed pallets with random distances and heights
        \item gaps
        \item grid stones - randomly placed stepping stones with random height differences
        \item beams - beams radiating outward from the center
    \end{itemize}
    \item \textbf{fine-tuning terrains} - terrains for stage 2 fine-tuning
    \begin{itemize}
        \item pentagon stones - randomly placed stones with pentagon surfaces.
        \item single-column stones - stepping stones in one column with height differences
        \item narrow pallets - pallets with smaller widths
        \item consecutive gaps
        \item narrow stairs
    \end{itemize}
\end{itemize} 
The visualization of the training terrains with curriculum can be found in movie 1 under the method section.

\subsubsection*{PPO parameters}
The PPO hyperparameters are listed in Table~\ref{tab:ppo}. All policies were trained with the same parameters for both ANYmal-D and GR-1.
\begin{table}
\centering
\caption{\textbf{PPO hyperparameters}. $4096$ parallelized environments. All policies were trained with the same parameters.}
\begin{tabular}{r|l}
parameter type & number \\ \hline 
batch size & $24 \cdot 4096 = 98304$ \\
mini batch size & $8 \cdot 4096 = 32768$ \\
number of epochs & $5$ \\
clip range & $0.2$ \\
entropy coefficient & $0.005$(stage 1) $0.002$(stage 2) \\
discount factor & $0.99$ \\
GAE discount factor & $0.95$ \\
desired KL-divergence & $0.01$ \\
learning rate & adaptive \\
\end{tabular}
\label{tab:ppo}
\end{table}

\subsubsection*{Joint Indices}
We use the following joint indices during training. 
\newline

\noindent ANYmal-D: The joint indices from 0 to 11 respectively stand for LF\_HAA, LF\_HFE, LF\_KFE, LH\_HAA, LH\_HFE, LH\_KFE, RF\_HAA, RF\_HFE, RF\_KFE, RH\_HAA, RH\_HFE, RH\_KFE. The detailed definition can be found in \cite{Hutter2016}.
\newline

\noindent GR-1: The joint indices from 0 to 22 respectively stand for l\_hip\_roll, l\_hip\_pitch, l\_knee\_pitch, l\_ankle\_pitch, l\_ankle\_roll, r\_hip\_roll, r\_hip\_yaw, r\_hip\_pitch, r\_knee\_pitch, r\_ankle\_pitch, r\_ankle\_roll, waist\_yaw, waist\_pitch, waist\_roll, l\_shoulder\_pitch, l\_shoulder\_roll, l\_shoulder\_yaw, l\_elbow\_pitch, r\_shoulder\_pitch, r\_shoulder\_roll, r\_shoulder\_yaw, r\_elbow\_pitch.

\subsection*{Ablation Study Details}
\subsubsection*{Network structures}
\begin{figure}
\centering
\includegraphics[width=0.5\textwidth]{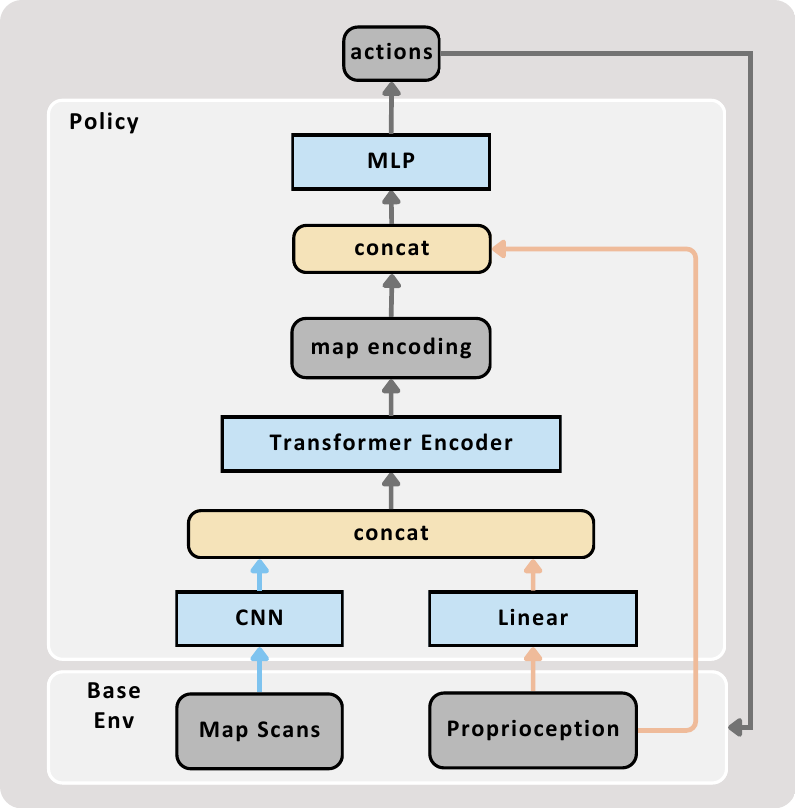}
\caption{\textbf{\highlight{Ablation Study on Network Structures - Transformer Encoder.}} We adopted the network structure proposed in~\cite{yang2022learning}. The map scans are embedded with a CNN and the proprioception is embedded with a linear layer. The embedded map and proprioception features are then concatenated and sent to a transformer encoder to generate a map encoding.}
\label{fig:ablation_trm}
\end{figure}

\begin{figure}
\centering
\includegraphics[width=0.5\textwidth]{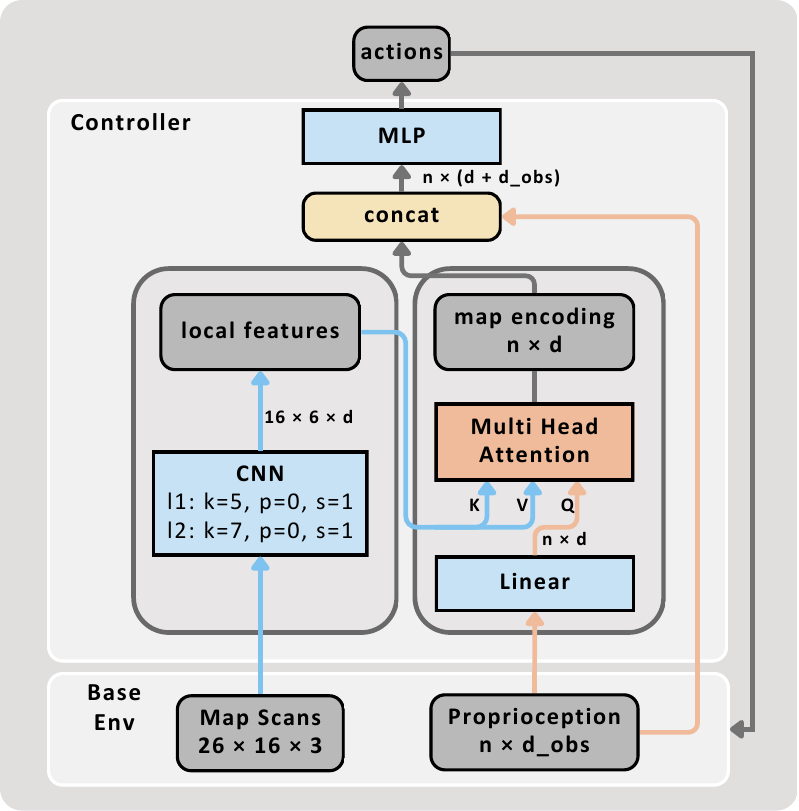}
\caption{\textbf{\highlight{Ablation Study on Network Structures - CNN down-sampling.}} The CNN consists of two layers with kernel sizes (k) of 5 and 7, respectively. The padding and stride are set to be 0 and 1 for both layers.}
\label{fig:ablation_CNN}
\end{figure}

\begin{figure}
\centering
\includegraphics[width=0.5\textwidth]{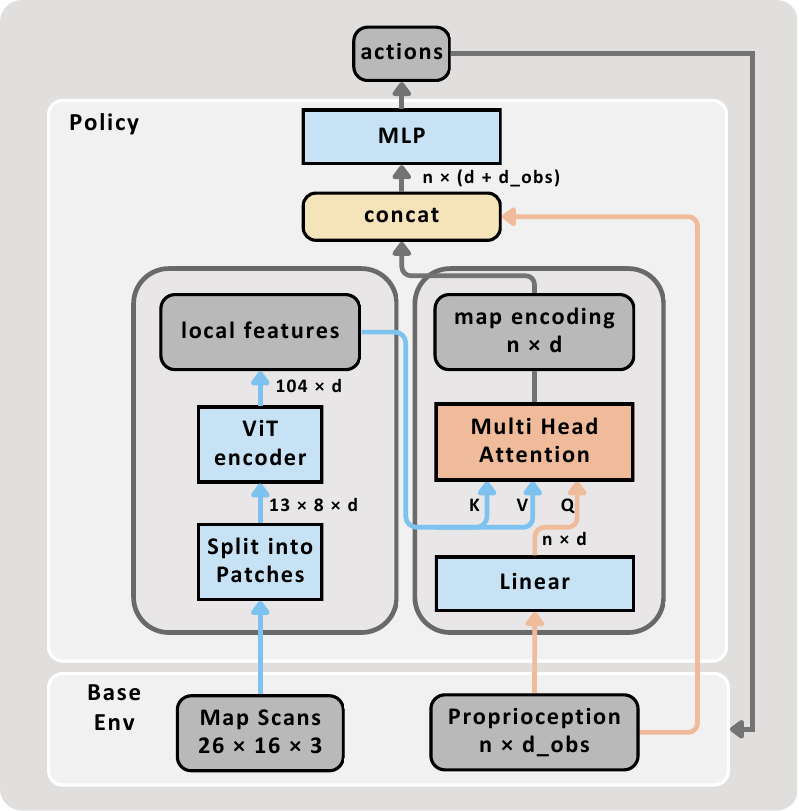}
\caption{\textbf{\highlight{Ablation Study on Network Structures - Vision Transformer.}} The map scans are split into 2 $\times$ 2 patches and sent to a vision transformer for feature extraction.}
\label{fig:ablation_vit}
\end{figure}

Figure~\ref{fig:ablation_trm}, ~\ref{fig:ablation_CNN}, and ~\ref{fig:ablation_vit} present the network structures used in the ablation study.


\end{document}